\documentclass{article}

\usepackage{microtype}
\usepackage{graphicx}
\usepackage{subfigure}
\usepackage{booktabs} % for professional tables

\usepackage{hyperref}

\usepackage[accepted]{icml2023}

\usepackage{amsmath}
\usepackage{amssymb}
\usepackage{mathtools}
\usepackage{amsthm}
\usepackage{algorithm}
\usepackage{algorithmic}
\usepackage{float} 
\usepackage{multirow} % Required for multirows
\usepackage{makecell}
\renewcommand{\algorithmicrequire}{ \textbf{Input:}} %Use Input in the format of Algorithm
\renewcommand{\algorithmicensure}{ \textbf{Output:}} %UseOutput in the format of Algorithm

\usepackage[capitalize,noabbrev]{cleveref}

\theoremstyle{plain}
\newtheorem{theorem}{Theorem}[section]

\newtheorem{lemma}[theorem]{Lemma}

\theoremstyle{definition}

\theoremstyle{remark}

\usepackage[textsize=tiny]{todonotes}

\icmltitlerunning{Submission and Formatting Instructions for ICML 2023}

\begin{document}

\twocolumn[
\icmltitle{Learning from Stochastic Labels}



\icmlsetsymbol{equal}{*}

\begin{icmlauthorlist}
\icmlauthor{Meng Wei }{equal,yyy}
\icmlauthor{Zhongnian Li}{equal,yyy}
\icmlauthor{Yong Zhou}{yyy}
\icmlauthor{Qiaoyu Guo}{yyy}
\icmlauthor{Xinzheng Xu}{yyy}

\end{icmlauthorlist}

\icmlaffiliation{yyy}{China University of Mining and Technology, Xuzhou, China}
%\icmlaffiliation{comp}{Company Name, Location, Country}
%\icmlaffiliation{sch}{School of ZZZ, Institute of WWW, Location, Country}

\icmlcorrespondingauthor{Xinzheng Xu}{xxzheng@cumt.edu.cn}

\icmlkeywords{Machine Learning, ICML}

\vskip 0.3in
]



\printAffiliationsAndNotice{\icmlEqualContribution} % otherwise use the standard text.

\begin{abstract}
Annotating multi-class instances is a crucial task in the field of machine learning. Unfortunately, identifying the correct class label from a long sequence of candidate labels is time-consuming and laborious. To alleviate this problem, we design a novel labeling mechanism called \emph{stochastic label}. In this setting, stochastic label  includes two cases: 1) identify a correct class label from a small number of randomly given labels; 2) annotate the instance with None label when given labels do not contain correct class label. In this paper, we propose a novel suitable approach to learn from these stochastic labels. We obtain an unbiased estimator that utilizes less supervised information in stochastic labels to train a multi-class classifier. Additionally, it is theoretically justifiable by deriving the estimation error bound of the proposed method. Finally, we conduct  extensive experiments on widely-used benchmark datasets to validate the superiority of our method by comparing it with existing state-of-the-art methods.

\end{abstract}

\section{Introduction}

\begin{figure*}[!htbp]
	\centering
	\includegraphics[width=5in]{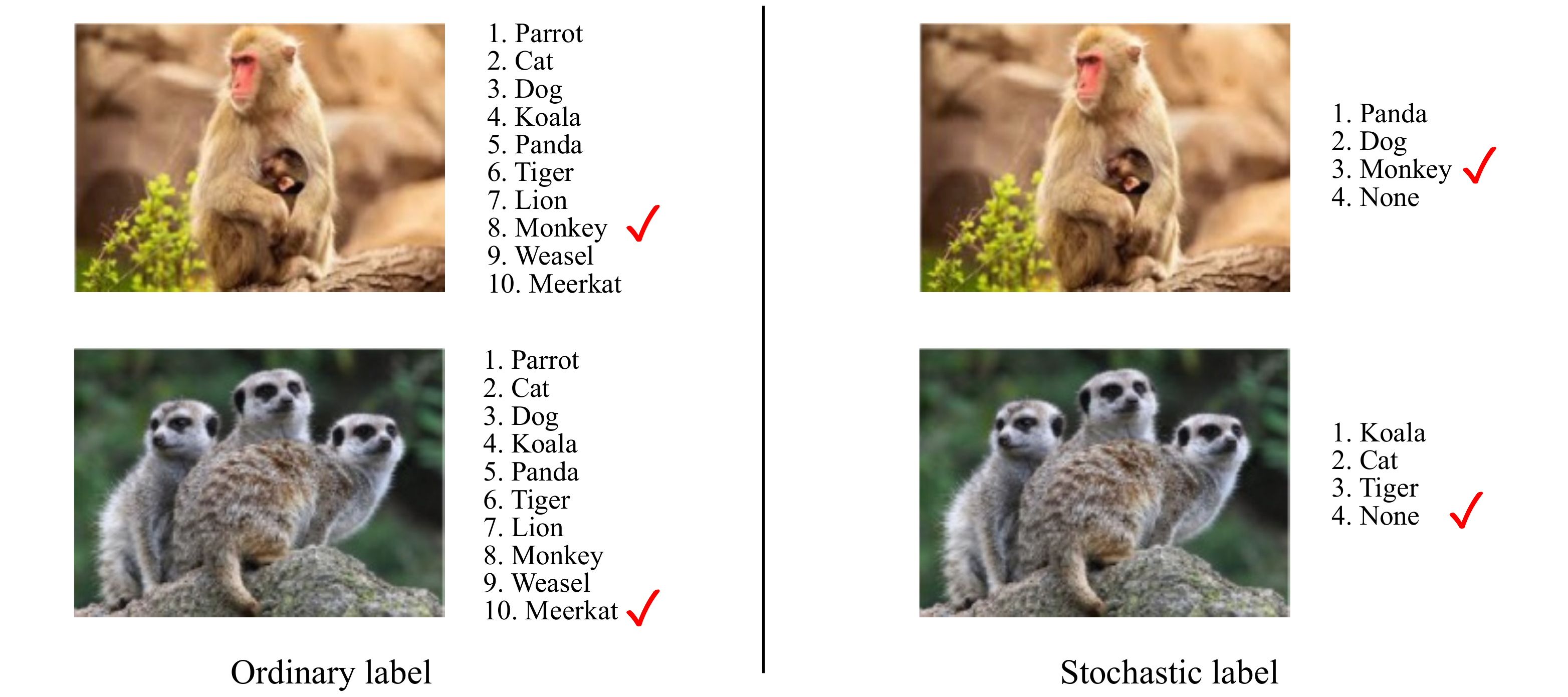}
	\caption{A comparison between ordinary label (left) and stochastic label (right). Here, the selected label is ticked. For the same instance, in ordinary label, crowdsourced workers need to identify the correct class label from 10 classes. However, in stochastic label, they only need to select from 3 stochastic labels: Dog, Monkey, Panda, or annotate None.}
\end{figure*}
Ordinary supervised classification needs to accurately select a ground-truth label for each instance from a long sequence of candidate labels, which is time-consuming and laborious in real-world datasets. To shorten the time and expense of labeling work, in the past decade, a large number of researchers have begun to study weakly supervised learning(WSL), hoping to learn an efficient classifier from weakly labels with less information. These WSL approaches include but not limited to, semi-supervised learning \citep{semi-supervised_1, semi-supervised_2, semi-supervised_3, semi-supervised_4, semi-supervised_5, semi-supervised_6}, noisy-label learning \citep{noisy_1, noisy_2, noisy_3, noisy_4, noisy_5, noisy_6, noisy_7}, partial-label learning \citep{pl_1, pl_2, pl_3, pl_4, pl_5, pl_6}, complementary-label learning  \citep{cll_1, cll_2, cll_3, cll_4, cll_5, cll_6, cll_7, cll_8} , unlabeled-unlabeled learning \citep{uu_1, uu_2} and positive-unlabeled learning \citep{pu_1, pu_2, pu_3, pu_4, pu_5, pu_6, pu_7}.

Most existing WSL works achieve satisfactory classification results in multi-class classification task. Nevertheless, these approaches all select one or more weakly labels from the entire candidate labels set, which inevitablely leads to consuming time and bad efficiency. Unfortunately, it is known that a long sequence of candidate labels is common in crowdsourcing Internet labeling jobs\citep{long_tail_1, long_tail_2}. Based on this fact, identifying the correct class label is a huge challenge. However, existing methods have not yet to reduce the size of the candidate labels during annotating instance.

To tackle this problem, stochastic label is proposed as a novel labeling mechanism to help annotators relieve from the heavy labeling tasks. Instead of precisely choosing the correct class label from the entire candidate labels set, stochastic label only needs to identify the ground-truth label from a given small number of labels or annotate None. For example, as shown in the right part of Figure 1, for a monkey instance, crowdsourced workers only need to identify the correct class label from given 3 labels. For a meerkat image, if the provided labels set does not contain the correct class label, the instance will be annotated with None. 

Collecting stochastic labels is undoubtedly easier and quicker than collecting ordinary labels. At the same time, this labeling mechanism is also very easy to implement and obtain in reality. As shown in Figure 1, for a monkey instance, crowdsourced workers usually need to browse through the entire candidate labels set, which is extremely time-consuming when the number of class is large. However, in stochastic label, for the same monkey instance, crowdsourced workers only need to find the monkey's label from 3 candidate labels, which takes much less time than choosing the ground-truth label from 10 class labels. Intutively, this case implies that accurately selecting the correct class label from stochastic labels set with 3 labels is preferable to that of the whole label set, which saves $ 70 \% $ of the time.

In this novel setting, we generate a small set of stochastic labels $ \tilde{Y} $ for each instance. In set $ \tilde{Y} $, each stochastic label is randomly and uniformly selected from the entire candidate labels set, i.e., $ \tilde{Y} \subset \{1, 2, \ldots, K\}$, and $ \forall i, j \in \tilde{Y}, P(i \in \tilde{Y}) = P(j \in \tilde{Y})$, where $ K $ denotes the number of classes and $ P(i \in \tilde{Y}) $ refers to the probability of selecting label $ i $ from the $ K $ classes. As mentioned earlier, for each instance, we either select a correct class label from the generated stochastic labels or annotate None. For these instances marked with None, we can utilize it as complementary labels. 

In order to utilize the less supervised information to train a multi-class classifier, we propose an unbiased risk estimator (URE) to learn from these stochastic labels. To stimulate more research on this problem, we establish a prototype baseline for this novel setting. Theoretically, we derive the upper bound of the evaluation risk of the proposed method, and prove that as the number of training samples increases, the empirical risk can converge to the real classification risk. To validate the usefulness of our proposed approach, we have done extensive experiments on widely-used benchmark datasets and compared it with state-of-the-out  methods.  

The rest of this paper is structured as follows. Section 2 gives a quick overview of related work. Section 3 presents the novel labeling mechanism and  the proposed approach with theoretical analyses. Section 4 reports the results of the comparative experiment. Finally, Section 5 concludes this paper. 

\section{Related Work}
In this section, we give notations and review the formulations of learning from ordinary labels, complementary labels, and partial labels. 

\subsection{Learning from Ordinary Labels}

Let $ \mathcal{X} \subset \mathbb{R}^{d}  $  be the feature space with $ d- $dimensions and $ \mathcal{Y} = \{1, 2, \ldots, K\} $ be the label space. $ y$ denotes the ground-truth class label of instance $  \textbf{\textit{x}} $. The sample $ (\textbf{\textit{x}}, y) $ is sampled from an unknown probability distribution with density $ p (\textbf{\textit{x}}, y) $. The aim of learning from ordinary labels is to train a classifier  $ f(\textbf{\textit{x}})\colon \textbf{\textit{x}} \mapsto \{1,2,\ldots,K\} $ that minimizes the expected risk:
\begin{equation}\label{eq1}
	R(f)=\mathbb{E}_{(\textbf{\textit{x}}, y)\sim p(\textbf{\textit{x}}, y)}[\ell(f(\textbf{\textit{x}}),y)], 
\end{equation}
where $\mathbb{E}$ refers to the expectation and $ \ell $ denotes the loss function. Suppose $  \mathcal{D}=\{(\textbf{\textit{x}}_{i}, y_{i}) \}_{i=1}^{N}  $ is a set of $ N $ training samples. The approximating empirical risk $ 	\hat{R}(f) $ can be defined as
\begin{equation}
	\hat{R}(f)=\frac{1}{N}\sum_{i=1}^{N}\ell(f(\textbf{\textit{x}}_i),y_i).
\end{equation}

\subsection{Learning from Complementary Label }
Different from ordinary label, complementary label specifies the class that the instance does not belong to. Suppose  $ \mathcal{\bar{Y}} = \{1,2,\ldots,K\} $ is the label space, and $(\textbf{\textit{x}}, \bar{y})\sim \bar{p}(\textbf{\textit{x}},\bar{y}) \neq p(\textbf{\textit{x}},y) $. Existing approaches on learning from complementary labels can be divided into two branches. 

The first branch assumes that the relationship between  $\bar{y}$ and $y$ is unbiased. As a pioneering work,  \citet{cll_1} proposed an unbiased risk estimator (URE) to train a classifier from single complementary label and provided solid theoretical analysis.  They assumed that  $  \bar{p}(\textbf{\textit{x}},\bar{y})  $ is defined as 
\begin{equation}
	\bar{p}(\textbf{\textit{x}},\bar{y})= \frac{1}{K-1} \sum_{y \neq \bar{y}} p(\textbf{\textit{x}}).
\end{equation}
However, this approach only works if the equation  $ \ell(z)+\ell(-z)=1 $ is satisfied. Accordingly, \citet{cll_3} proposed a general URE, which works for arbitrary loss function. In their paper, the expected risk can be expressed as 
\begin{equation}
	\begin{split}
		R(f) = \sum_{k=1}^{K} \bar{\pi}_{k} \mathbb{E}_{\bar{p}(\textbf{\textit{x}}, \bar{y})_k}[- (K - 1) \cdot \ell(k, f(\textbf{\textit{x}})) 
		\\ + \sum_{j=1}^{K} \ell(j, f(\textbf{\textit{x}}))],
	\end{split}
\end{equation}
where  $ \bar{\pi}_{k} $ refers to the proportion of the number of complementary labels of $k-$class and $ \mathbb{E}_{\bar{p}(\textbf{\textit{x}}, \bar{y})_k} $ denotes the expectation of  $k-$class. 

Different from unbiased assumption,  \citet{cll_2} assumed that the relationship is biased. They assumed that all other labels except the ground-truth label can be the complementary label with different probabilities. By using a transfer probabilities matrix, $ P(\bar{y} = j \mid  \textbf{\textit{x}}) $ can be written as 
\begin{equation}
	P(\bar{y} = j \mid  \textbf{\textit{x}}) = \sum_{i \neq j}P(\bar{y} = j \mid  y = i)P(y = i \mid  \textbf{\textit{x}}).
\end{equation}
However, they did not provide an unbiased risk estimator.

Unlike previous work, only one complementary label was used, \citet{cll_5} proposed a novel setting for multiple complementary labels, where each instance is equipped with a set of complementary labels. They assumed that the distribution of correct class label $ y $ and complementary labels set $ \bar{Y} $ can be written as
\begin{equation} \label{eq6}
	p(\textbf{\textit{x}}, \bar{Y} \mid  y \notin \bar{Y}) = \sum_{j=1}^{K-1} p(l=j)\bar{p}(\textbf{\textit{x}}, \bar{Y} \mid  l = j),   
\end{equation} 
where $ \bar{Y} $ denotes the complementary labels set, and $ l $ refers to the size of complementary labels set.  By using  equation (\ref{eq6}), \citet{cll_5} proposed an unbiased risk estimator to learn from multiple complementary labels, with treating the multiple complementary labels as a whole.

\subsection{Learning from Partial Labels}
In partial label learning, each training instance is provided with a set of candidate labels, where  the ground-truth label exists in this set of candidate labels. The opposite of this partial labels set is the complementary labels. The remaining labels can be used as multiple complementary labels after removing the partial labels. Thus, learning from partial labels can be seen as another approach to solve the problem of complementary labels. Although the current stage of partial label learning has achieved great success, and \citet{pl_7} have provided proof of statistical consistency, no one has yet given an unbiased estimator of the classification risk.

All of the aforementioned methods are chosen from a long sequence of candidate labels. In this paper, we propose a novel labeling mechanism where selecting a label from a given small number of labels or annotating None.

\section{Learning from Stochastic Labels}
In this section, we propose a novel labeling mechanism called stochastic label and provide an unbiased risk estimator framework to learn from the stochastic label.

\subsection{Notation}
We consider another scenario where each instance is equipped with a small set of stochastic labels $ \tilde{Y} $  instead of ordinary label $ y $. Suppose $ \textbf{\textit{x}} \in \mathbb{R}^{d} $ is the the $ d- $dimensional feature vector and $ \tilde{Y}  \subseteq \{1, 2, \ldots, K\} $ denotes a stochastic labels set for instance $ \textbf{\textit{x}} $. Suppose $ s = j $ indicates that $ \tilde{Y} $ contains ground-truth label $ j $, and $ s = K + 1 $ denotes that the provided $ \tilde{Y} $ does not contain ground-truth label, where $   j \in \{1, 2, \ldots,  K\} $ and $ K $ denotes the size of whole candidate labels set. Suppose $ s $ is sampled from $ P(s) $. Let $ \tilde{D}  = \{(\textbf{\textit{x}}_i, \tilde{Y}_i, s_i) \}_{i=1}^{N}$ be sampled randomly and uniformly from an unknown probability distribution with density $ P(\textbf{\textit{x}}, \tilde{Y}, s) $, where $ N $ denotes the size of training samples. For ease of reading, let $ M $ denotes $ P(\textbf{\textit{x}}, \tilde{Y}, s) $. Let $ M_{+} $ denotes $ P(\textbf{\textit{x}}, \tilde{Y} \mid  s = j)$ and $ M_{-} $ denotes $ P(\textbf{\textit{x}}, \tilde{Y} \mid  s = K + 1) $. Let $ l $ be the size of the stochastic labels set. Let $ \mathcal{L} $ be the multi-class loss function. Our goal is to learn a multi-class classifier $ f: \textbf{\textit{x}} \mapsto \{1, 2, \dots, K\} $ from these stochastic labels to minimize classification risk.

\subsection{Labeling Mechanism}
Each stochastic label from $ \tilde{Y} $ is randomly and uniformly selected from the entire candidate labels set. The probability of selecting label $ j $ from  $ K $ classes can be expressed as

\begin{equation}
	P(j \in \tilde{Y} \mid l) = \frac{l}{K}.
\end{equation}
It is obvious that the conditional probabilities of $ y $ and $ \tilde{Y} $ satisfy the following property when $ s = K + 1 $: 
\begin{equation}\label{eq8}
	P(y=j, \tilde{Y} \mid  \textbf{\textit{x}}, s = K + 1) = \frac{1}{K-l}P(\tilde{Y} \mid \textbf{\textit{x}}, s = K + 1).
\end{equation}

Specifically, if $ l = K-1 $ , the problem of learning from ordinary labels will be obtained. The expirical approximation of the $ P(s) $ can be expressed by following equation:
\begin{equation}\label{eq7}
	P(s=z) = \frac{1}{N} \sum_{i=1}^{N} \mathbb{I}(s_i = z),
\end{equation}
where $ \mathbb{I}(\cdot) $ is the indicator function and $ z \in \{1, 2, \dots, K+1\} $. To sum up, in the labeling stage, some stochastic labels are randomly selected for each instance, and $ s $ is determined based on whether the ground-truth label exists  in these stochastic labels.

\begin{algorithm}[htbp]
	\caption{ Learning from Stochastic Labels}
	\label{alg:Framwork}
	\begin{algorithmic}
		\REQUIRE ~~\\ 
		$ \tilde{D}  = \{(\textbf{\textit{x}}_i, \tilde{Y}_i, s_i) \}_{i=1}^{N}$ is sampled randomly and uniformly from $ P(\textbf{\textit{x}}, \tilde{Y}, s) $;\\
		The number of epochs, $ T $;\\
		An external stochastic optimization algorithm, $ \mathcal{A} $; 
		\FOR{$t=1$ to $T$}
		\STATE Calculate $ P(s) $ according equation (\ref{eq7});
		\STATE Shuffle $ \{(\textbf{\textit{x}}_i, \tilde{Y}_i, s_i)\} $ into $ B $ mini-batches with size $ n $;
		\FOR{$ b=1 $ to $ B $}
		\STATE Let $ N_s $ be the number of $ s= K + 1 $;
		\FOR{$ i=1  $ to $ n $}
		\IF{$ s_i = K + 1 $}
		\STATE Calculate $ \tilde{\mathcal{L}}[f(\textbf{\textit{x}}), \tilde{Y}] $;
		\STATE Denotes $ r_{i}(\theta) = \frac{1}{N_s} P(s_i=K+1)  \frac{1}{K-l} \sum_{j \notin \tilde{Y}_i}\mathcal{L}[f(\textbf{\textit{x}}_i), j]  $;
		\ELSE
		\STATE Calculate $ \mathcal{L}[f(\textbf{\textit{x}}), s_i] $;
		\STATE Denotes $ r_{i}(\theta)= \frac{1}{N-N_s} P(s_i) \mathcal{L}[f(\textbf{\textit{x}}_i), s_i] $;
		\ENDIF
		\ENDFOR
		\STATE Denotes $ L^{b}(\theta)=\sum_{i=1}^{n}r_{i}(\theta) $;
		\STATE Set gradient $ -\bigtriangledown_{\theta}L^{b}(\theta) $;
		\STATE Update $ \theta $ by $ \mathcal{A} $;
		\ENDFOR
		\ENDFOR
		\ENSURE Model parameter $ \theta $ for $f(\textbf{\textit{x}}, \theta)$;
	\end{algorithmic}
\end{algorithm}

\begin{figure*}[!htbp]
	\centering
	% mnist
	\subfigure[MNIST, size=8]{
		\begin{minipage}[b]{0.25\linewidth}
			\centering
			\includegraphics[width=2.0in]{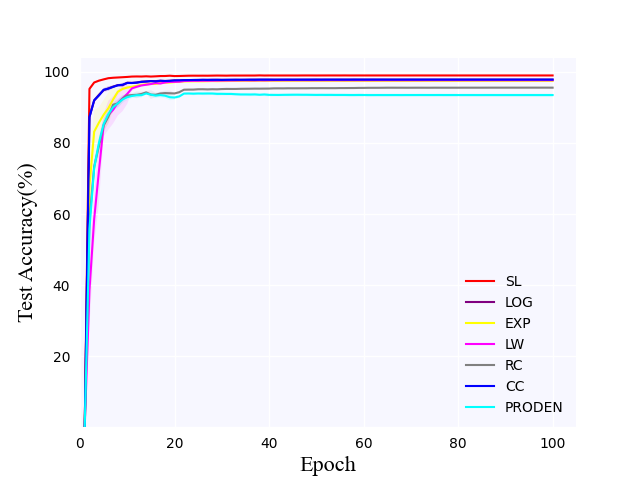}
		\end{minipage}
	}%
	\subfigure[MNIST, size=7]{
		\begin{minipage}[b]{0.25\linewidth}
			\centering
			\includegraphics[width=2.0in]{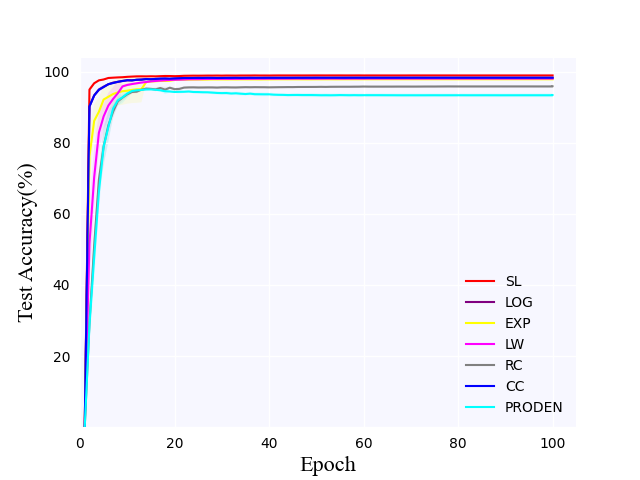}
			%\caption*{}
		\end{minipage}
	}%
	\subfigure[MNIST, size=6]{
		\begin{minipage}[b]{0.25\linewidth}
			\centering
			\includegraphics[width=2.0in]{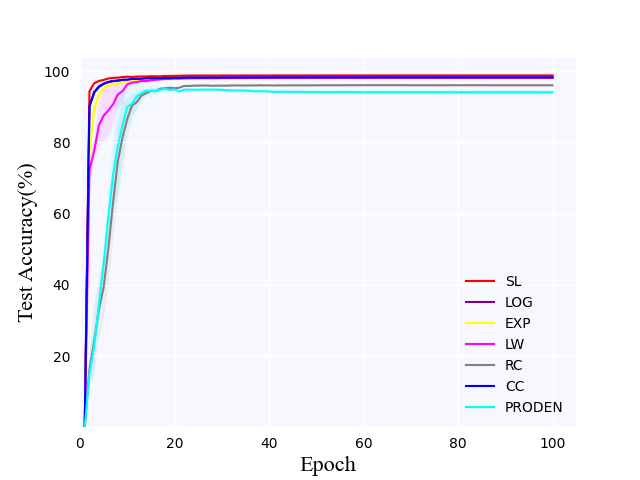}
			%\caption*{}
		\end{minipage}
	}%
	\subfigure[MNIST, size=5]{
		\begin{minipage}[b]{0.25\linewidth}
			\centering
			\includegraphics[width=2.0in]{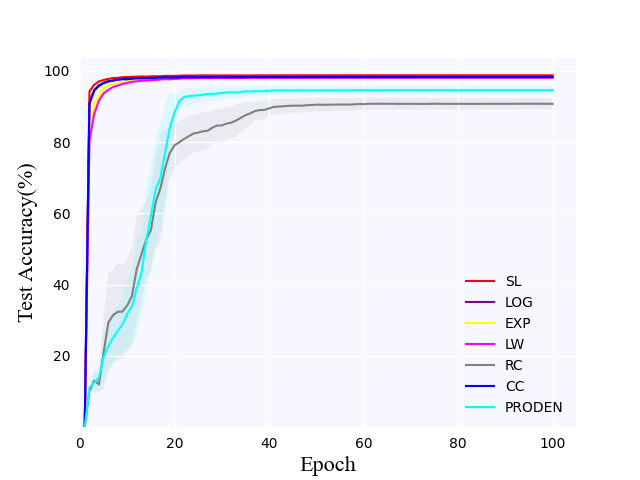}
			%\caption*{}
		\end{minipage}
	}%
	
	%cifar10
	\subfigure[CIFAR10, size=8]{
		\begin{minipage}[b]{0.25\linewidth}
			\centering
			\includegraphics[width=2.0in]{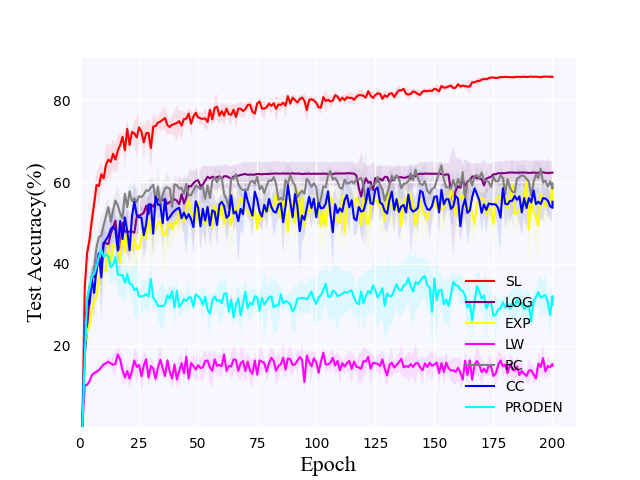}
			%\caption*{}
		\end{minipage}
	}%
	\subfigure[CIFAR10, size=7]{
		\begin{minipage}[b]{0.25\linewidth}
			\centering
			\includegraphics[width=2.0in]{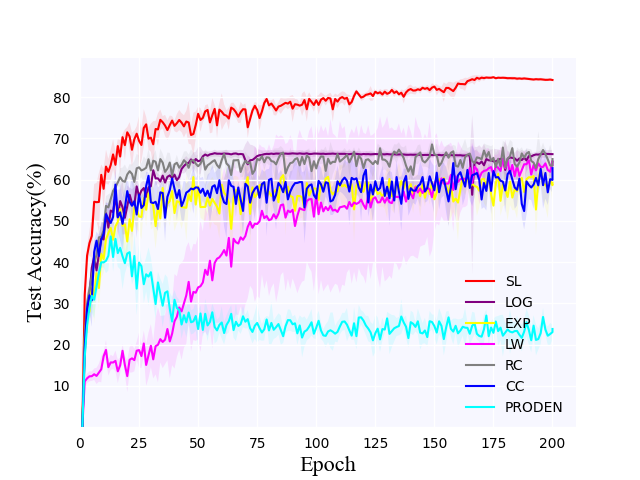}
			%\caption*{}
		\end{minipage}
	}%
	\subfigure[CIFAR10, size=6]{
		\begin{minipage}[b]{0.25\linewidth}
			\centering
			\includegraphics[width=2.0in]{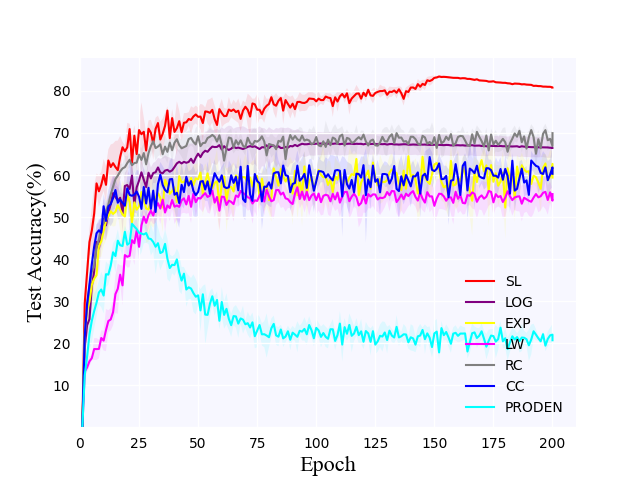}
			%\caption*{}
		\end{minipage}
	}%
	\subfigure[CIFAR10, size=5]{
		\begin{minipage}[b]{0.25\linewidth}
			\centering
			\includegraphics[width=2.0in]{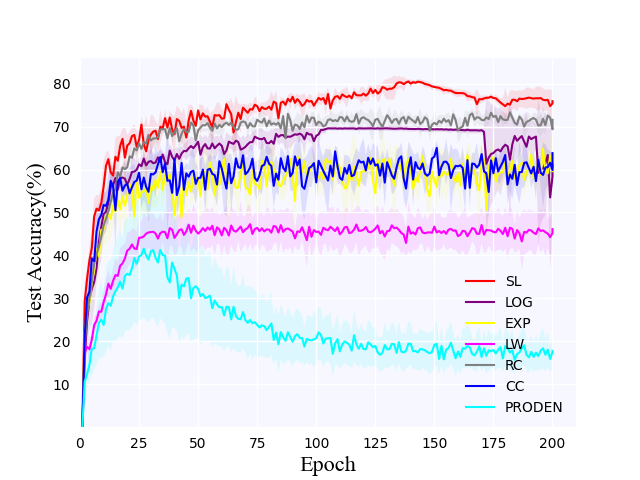}
			%\caption*{}
		\end{minipage}
	}%
	\caption{ Experiments results of test classification accuracy of various datasets. The dark colors show the mean accuracy of 5 trials and the light colors show the standard deviation.}
\end{figure*}

\subsection{Unbiased Risk Estimator}

We start by considering the case where $ s = K + 1 $. According equation (\ref{eq8}),   the following theorem can be obtained.
\begin{theorem}
	For any instance $ \textbf{\textit{x}} $ with its ground-truth label $ y $ and stochastic labels set $ \tilde{Y} $, the following equality holds:
	\begin{equation}
		\begin{split}
			P(j \mid \textbf{\textit{x}}) = P(y=j \mid \textbf{\textit{x}}, s = j)P(s = j) 
			\\  + \frac{1}{K-l}\sum_{j \notin \tilde{Y}}P(\tilde{Y} \mid \textbf{\textit{x}}, s = K + 1) P(s=K+1).
		\end{split}
	\end{equation}
\end{theorem}
The proof is presented in Appendix. Let us consider the loss of the stochastic labels. By using Theorem 3.1, we have the following equation when  $ s = K + 1 $:
\begin{equation}\label{eq10}
	\begin{split}
		\tilde{\mathcal{L}}[f(\textbf{\textit{x}}), \tilde{Y}] =  \frac{1}{K-l} \sum_{j \notin \tilde{Y}}\mathcal{L}[f(\textbf{\textit{x}}), j].
	\end{split}
\end{equation}

Based on Theorem 3.1 and equation (\ref{eq10}), an unbiased risk estimator of learning from stochastic labels can be derived by the following theorem.
\begin{theorem}
	The classification risk (\ref{eq1}) can be described as
	\begin{equation}\label{eq11}
		\begin{split}
			R(f) & = \mathbb{E}_{M_{+}}P(s=j)\mathcal{L}[f(\textbf{\textit{x}}), j] \\
			& \qquad + \mathbb{E}_{M_{-}} P(s=K+1) \tilde{\mathcal{L}}[f(\textbf{\textit{x}}, \tilde{Y} )],
		\end{split}
	\end{equation}
\end{theorem}
where 
\begin{equation}
	\begin{aligned}
		\mathbb{E}_{M_{+}} & = \mathbb{E}_{M} \sum_{j} P(y=j \mid \textbf{\textit{x}}, s=j) \\
		& = \mathbb{E}_{(\textbf{\textit{x}}, j) \sim P(\textbf{\textit{x}}, j \mid s = j)}
	\end{aligned}
	\nonumber
\end{equation}
when $  s = j $, and 
\begin{equation}
	\begin{aligned}
		\mathbb{E}_{M_{-}} & = \mathbb{E}_{M}P(\tilde{Y} \mid \textbf{\textit{x}}, s=K+1) \\
		& = \mathbb{E}_{(\textbf{\textit{x}}, \tilde{Y}) \sim P(\textbf{\textit{x}}, \tilde{Y} \mid s = K+1)}
	\end{aligned}
	\nonumber
\end{equation}
when $ s = K + 1 $.

The proof is presented in Appendix. Since the training dataset $ \tilde{D} =  \{(\textbf{\textit{x}}_i, \tilde{Y}_i, s_i) \}_{i=1}^{N}$ are sampled randomly and uniformly from $ P(\textbf{\textit{x}}, \tilde{Y}, s) $, the expression (\ref{eq11}) can be naively approximated by 
\begin{equation}\label{eq12}
	\begin{split}
		& \hat{R}(f) = \frac{1}{N-N_s} \sum_{i=1}^{N-N_s} P(s=j) \mathcal{L}[f(\textbf{\textit{x}}_i), j]  \\
		& +  \frac{1}{N_s} \sum_{i=1}^{N_s} P(s=K+1) \frac{1}{K-l}  \sum_{j  \notin \tilde{Y}_i}\mathcal{L}[f(\textbf{\textit{x}}_i), j], 
	\end{split}
\end{equation}
where $ N_s $ denotes the size of samples when $ s = K + 1 $.

In the training phase, we use stochastic gradient descent to optimize expression (\ref{eq12}). When given a set of training samples, we calculate its loss based on whether it contains real labels. The detail of the overall algorithm procedure of the proposed approach is shown in Algorithm 1.

\subsection{Estimation Error Bound}
Here, we analyze the generalization estimation error bound for the proposed URE. Let $ f $ be the classification vector function in the hypothesis set $ \mathcal{F} $. Using  $ \varphi_{\mathcal{L}} $ and $ p^{*} $ to denote the upper bound of the loss function $ \mathcal{L} $ and $ P(s) $, i.e., $ \mathcal{L}[f(\textbf{\textit{x}}_{i}, y_{i})] \leqslant \varphi_{\mathcal{L}} $ and $ P(s = j) \leqslant p^{*}, j \in \{1, \ldots, K+1\} $. Using $ L_{f} $ and $ \mathfrak{R}_{N}(\mathcal{F}) $  to denote the Lipschitz constant \citep{rademacher} of $ f $ and the Rademacher complexities \citep{rademacher} of $ \mathcal{F} $, we can establish the following lemma.

\begin{lemma}
	For any $ \delta > 0 $, with the probability at least $ 1- \delta / 2 $, we have 
	\begin{equation}
		\begin{split}
			\mathop{sup}_{f\in\mathcal{F}} \mid  \hat{R}_{OL}(f) - R_{OL}(f) \mid  \leqslant 2 p^{*}L_{f} \mathfrak{R}_{N}(\mathcal{F}) \\
			+ (p^{*} \varphi_{\mathcal{L}})\sqrt{\frac{2log\frac{4}{\delta}}{N}},
		\end{split}
	\end{equation}
	and
	\begin{equation}
		\begin{split}
			\mathop{sup}_{f\in\mathcal{F}} \mid  \hat{R}_{CL}(f) - R_{CL}(f) \mid  \leqslant 2 p^{*}(K-l)L_{f} \mathfrak{R}_{N}(\mathcal{F}) \\
			+ p^{*} (K-l)\varphi_{\mathcal{L}}\sqrt{\frac{2log\frac{4}{\delta}}{N}},
		\end{split}
	\end{equation}
\end{lemma}

where $ R_{OL}(f) = \mathbb{E}_{M_{+}} $, $ R_{CL}(f) = \mathbb{E}_{M_{-}} $,  $ \hat{R}_{OL}(f) $ and $ \hat{R}_{CL}(f)  $  denotes the empirical risk  of $ R_{OL}(f) $ and $ R_{CL}(f) $. The proof is given in Appendix.

Based on Lemma 3.3, the estimation error bound can be expressed as follows.
\begin{theorem}
	For any $ \delta > 0 $, with the probability at least $ 1- \delta / 2 $, we have 
	\begin{equation}
		\begin{split}
			R(\hat{f}) - R(f^{\ast})  \leqslant 4 p^{*}(K-l+1)L_f \mathfrak{R}_{N}(\mathcal{F}) \\
			+ 2 p^{*} \varphi_{\mathcal{L}} \sqrt{\frac{2log\frac{4}{\delta}}{N}} +  2p^{*} (K-l)\varphi_{\mathcal{L}} \sqrt{\frac{2log\frac{4}{\delta}}{N}},
		\end{split}
	\end{equation}
\end{theorem}
where $ \hat{f} $ denotes the trained classifier, $ R(f^{\ast}) = \underset{f\in \mathcal{F}}{min}R(f)  $. The proof is presented in Appendix.

Lemma 3.3 and Theorem 3.4 show that our method exists an estimation error bound. With the deep network hypothesis set $ \mathcal{F} $ fixed, we have $ \mathfrak{R}_{N}(\mathcal{F}) = \mathcal{O}(1/\sqrt{N}) $. Therefore, with  $ N \longrightarrow \infty $, we have $R(\hat{f}) = R(f^{\ast}) $, which proves that our method could converge to the optimal solution. 

\begin{table*}[!htbp]
	\renewcommand{\arraystretch}{1}
	\caption{Test classification accuracy (mean $ \pm $ std $ \% $) of 5 trials on various datasets and approaches. The data at hand is from all class. The best performance is shown in bold.}
	\label{table_2}
	\centering
	\begin{tabular}{c|c|c|c|c|c|c|c}%c表示文本居中，c的个数是列数
		\toprule[1pt] 
		Dataset & Method  &  size = 8  & size = 7  & size = 6  & size = 5  & size = 4  & size = 3  \\
		\midrule
		\multirow{7}{*}{MNIST}
		& SL (our) & \textbf{99.04 $\pm$ 0.06} & \textbf{98.95 $\pm$ 0.05} & \textbf{98.86 $\pm$ 0.04} & \textbf{98.81 $\pm$ 0.02} & \textbf{98.70 $\pm$ 0.04} & \textbf{98.55 $\pm$ 0.12}  \\
		& LOG & 97.86 $\pm$ 0.17 & 98.28 $\pm$ 0.12 & 98.30 $\pm$ 0.03 & 98.38 $\pm$ 0.08 & 98.24 $\pm$ 0.08 & 98.18 $\pm$ 0.11  \\
		& EXP & 97.54 $\pm$ 0.16 & 97.87 $\pm$ 0.15 & 98.09 $\pm$ 0.08 & 98.22 $\pm$ 0.13 & 98.07 $\pm$ 0.10 & 97.99 $\pm$ 0.10 \\
		& LW & 97.70 $\pm$ 0.09 & 98.06 $\pm$ 0.05 & 98.19 $\pm$ 0.12 & 98.14 $\pm$ 0.10 & 95.21 $\pm$ 2.34 & 40.48 $\pm$ 25.26 \\
		& RC & 95.60 $\pm$ 0.18 & 95.84 $\pm$ 0.28 & 96.12 $\pm$ 0.21 & 90.08 $\pm$ 1.64 & 21.01 $\pm$ 12.33 & 10.52 $\pm$ 0.70 \\
		& CC & 97.85 $\pm$ 0.11 & 98.29 $\pm$ 0.13 & 98.28 $\pm$ 0.03 & 98.33 $\pm$ 0.08 & 98.16 $\pm$ 0.10 & 98.12 $\pm$ 0.70 \\
		& PRODEN & 94.23 $\pm$ 0.36 & 95.22 $\pm$ 0.25 & 95.54 $\pm$ 0.15 & 94.67 $\pm$ 1.18 & 13.41 $\pm$ 4.1 & 11.35 $\pm$ 0.00 \\
		\midrule
		\multirow{7}{*}{Fashion}
		& SL (our) & \textbf{91.64 $\pm$ 0.25} & \textbf{91.43 $\pm$ 0.17} & \textbf{90.93 $\pm$ 0.25} & \textbf{90.89 $\pm$ 0.26} & \textbf{90.49 $\pm$ 0.23} & \textbf{89.68 $\pm$ 0.20} \\
		& LOG & 88.80 $\pm$ 0.22 & 89.10 $\pm$ 0.19 & 89.58 $\pm$ 0.29 & 89.35 $\pm$ 0.14 & 89.15 $\pm$ 0.24 & 88.83 $\pm$ 0.18 \\
		& EXP & 87.65 $\pm$ 0.33 & 88.28 $\pm$ 0.29 & 88.58 $\pm$ 0.23 & 88.49 $\pm$ 0.23 & 88.38 $\pm$ 0.21 & 88.03 $\pm$ 0.37 \\
		& LW & 88.53 $\pm$ 0.25 & 88.81 $\pm$ 0.21 & 88.82 $\pm$ 0.46 & 87.15 $\pm$ 3.00 & 83.00 $\pm$ 4.21 & 58.06 $\pm$ 13.22\\
		& RC & 86.20 $\pm$ 0.15 & 85.58 $\pm$ 0.36 & 85.60 $\pm$ 0.49 & 81.37 $\pm$ 2.44 & 31.48 $\pm$ 4.94 & 10.21 $\pm$ 0.24 \\
		& CC & 88.94 $\pm$ 0.25 & 89.20 $\pm$ 0.25 & 89.70 $\pm$ 0.22 & 89.50 $\pm$ 0.32 & 89.22 $\pm$ 0.22 & 88.91 $\pm$ 0.20 \\
		& PRODEN & 85.03 $\pm$ 0.33 & 84.90 $\pm$ 0.51 & 83.89 $\pm$ 0.35 & 82.33 $\pm$ 1.06 & 47.82 $\pm$ 26.11 & 10.02 $\pm$ 0.01 \\
		\midrule
		\multirow{7}{*}{Kuzushiji}
		& SL (our) & \textbf{94.43 $\pm$ 0.12} & \textbf{94.19 $\pm$ 0.07} & \textbf{93.87 $\pm$ 0.21} & \textbf{93.15 $\pm$ 0.08} & \textbf{92.43 $\pm$ 0.17} & \textbf{91.60 $\pm$ 0.33}\\
		& LOG & 88.33 $\pm$ 0.39 & 89.43 $\pm$ 0.38 & 89.77 $\pm$ 0.26 & 89.35 $\pm$ 0.31 & 88.99 $\pm$ 0.44 & 87.93 $\pm$ 0.61 \\
		& EXP & 83.20 $\pm$ 3.12 & 86.64 $\pm$ 0.50 & 86.48 $\pm$ 1.85 & 86.09 $\pm$ 1.39 & 86.37 $\pm$ 0.32 & 85.11 $\pm$ 1.51 \\
		& LW & 83.36 $\pm$ 1.29 & 85.58 $\pm$ 1.22 & 84.77 $\pm$ 2.92 & 83.79 $\pm$ 1.27 & 29.88 $\pm$ 10.15 & 19.85 $\pm$ 4.48 \\
		& RC & 78.35 $\pm$ 0.64 & 76.63 $\pm$ 0.46 & 75.59 $\pm$ 2.04 & 51.62 $\pm$ 0.06 & 11.16 $\pm$ 1.73 & 10.01 $\pm$ 0.02 \\
		& CC & 88.35 $\pm$ 0.36 & 89.26 $\pm$ 0.41 & 89.60 $\pm$ 0.41 & 89.21 $\pm$ 0.32 & 88.78 $\pm$ 0.33 & 87.80 $\pm$ 0.34 \\
		& PRODEN & 75.63 $\pm$ 1.18 & 75.27 $\pm$ 1.10 & 74.75 $\pm$ 1.73 & 63.33 $\pm$ 3.26 & 10.06 $\pm$ 0.05 & 10.02 $\pm$ 0.01 \\
		\midrule
		\multirow{7}{*}{CIFAR10}
		& SL (our) & \textbf{86.06 $\pm$ 0.26} & \textbf{84.95 $\pm$ 0.32} & \textbf{83.61 $\pm$ 0.21} & \textbf{81.40 $\pm$ 0.42} & \textbf{79.05 $\pm$ 0.38} & 74.58 $\pm$ 0.52 \\
		& LOG & 62.69 $\pm$ 3.00 & 66.90 $\pm$ 0.85 & 67.76 $\pm$ 2.69 & 69.96 $\pm$ 0.36 & 68.64 $\pm$ 1.56 & 66.43 $\pm$ 1.63 \\
		& EXP & 62.87 $\pm$ 0.93 & 66.94 $\pm$ 0.46 & 68.23 $\pm$ 0.19 & 68.32 $\pm$ 0.76 & 67.36 $\pm$ 1.26 & 65.43 $\pm$ 1.35 \\
		& LW & 22.89 $\pm$ 1.55 & 67.62 $\pm$ 1.41 & 58.46 $\pm$ 2.86 & 49.02 $\pm$ 4.45 & 28.06 $\pm$ 6.55 & 14.18 $\pm$ 2.41 \\
		& RC & 66.09 $\pm$ 0.55 & 69.99 $\pm$ 0.47 & 72.57 $\pm$ 0.87 & 74.87 $\pm$ 0.78 & 75.93 $\pm$ 0.37 & \textbf{75.92 $\pm$ 0.43} \\
		& CC & 63.02 $\pm$ 0.88 & 67.57 $\pm$ 0.44 & 67.75 $\pm$ 0.49 & 68.60 $\pm$ 0.79 & 67.81 $\pm$ 0.69 & 66.31 $\pm$ 0.39 \\
		& PRODEN & 45.87 $\pm$ 1.36 & 47.83 $\pm$ 2.59 & 50.78 $\pm$ 2.67 & 52.07 $\pm$ 1.49 & 52.39 $\pm$ 1.30 & 47.96 $\pm$ 1.66 \\
		\bottomrule[1.5pt]
	\end{tabular}
\end{table*}

\begin{table*}[!htbp]
	\renewcommand{\arraystretch}{1}
	\caption{Test classification accuracy (mean $ \pm $ std $ \% $) of 5 trials using stochastic label(SL) and ordinary label(OL) on various datasets.}
	\label{table_1}
	\centering
	\begin{tabular}{c|c|c|c|c|c|c|c}%c表示文本居中，c的个数是列数
		\toprule[1pt] 
		Method & class & size & proportion & MNIST & Fashion & Kuzushiji & CIFAR10  \\
		\midrule
		\multirow{4}{*}{SL(our)}
		& 1 $ \sim $ 10 & 8 & 0.2  & \textbf{99.04$ \pm $ 0.06} & \textbf{91.64 $ \pm $ 0.25} & \textbf{94.43 $ \pm $ 0.12} & \textbf{86.06 $ \pm $ 0.26}\\
		& 1 $ \sim $ 10  & 7 & 0.3  & 98.95$ \pm $ 0.05 & \textbf{ 91.43 $ \pm $ 0.17 } & \textbf{94.19 $ \pm $ 0.07} & \textbf{84.95 $ \pm $ 0.32} \\
		& 1 $ \sim $ 10 & 6 & 0.4  & 98.86$ \pm $ 0.04 & 90.93 $ \pm $ 0.25 & 93.87 $ \pm $ 0.21 & \textbf{83.61 $ \pm $ 0.21}\\
		& 1 $ \sim $ 10 & 5 & 0.5  & 98.81$ \pm $ 0.02 & 90.89 $ \pm $ 0.26 & 93.15 $ \pm $ 0.08 & 81.40 $ \pm $ 0.42\\
		\midrule
		OL & 1 $ \sim $ 10 & - & - & 98.97$ \pm $ 0.07 & 91.31 $ \pm $ 0.20 & 93.88 $ \pm $ 0.21 & 82.81 $ \pm $ 0.56\\
		\bottomrule[1pt]
	\end{tabular}
\end{table*}

\begin{table*}[!htbp]
	\renewcommand{\arraystretch}{1}
	\caption{Test classification accuracy (mean $ \pm $ std $ \% $) of 5 trials on various datasets and approaches.}
	\label{table_3}
	\centering
	\begin{tabular}{c|c|c|c|c|c|c|c|c|c}%c表示文本居中，c的个数是列数
		\toprule[1pt] 
		class & size & Dataset  & LOG & EXP & LW & RC & CC & PRODEN & SL (our)  \\
		\midrule
		1 $ \sim $ 8 & 4 & MNIST  & \makecell{98.66 \\ ($ \pm $ 0.10)} & \makecell{98.46 \\ ($ \pm $ 0.09)} & \makecell{98.55 \\ ($ \pm $ 0.08)} & \makecell{97.39 \\ ($ \pm $ 0.14)} & \makecell{98.65 \\ ($ \pm $ 0.06)} & \makecell{97.19 \\ ($ \pm $ 0.15)} &  \makecell{\textbf{99.20} \\ ($ \pm $ 0.06)} \\
		1 $ \sim $ 8 & 4 & Fashion  & \makecell{88.40 \\ ($ \pm $ 0.23)} & \makecell{87.65 \\ ($ \pm $ 0.35)} & \makecell{87.79 \\ ($ \pm $ 0.35)} & \makecell{84.18 \\ ($ \pm $ 0.69)} & \makecell{88.69 \\ ($ \pm $ 0.31)} & \makecell{81.51 \\ ($ \pm $ 0.96)} &  \makecell{\textbf{89.68} \\ ($ \pm $ 0.31)} \\
		1 $ \sim $ 8 & 4 & Kuzushiji  & \makecell{90.56 \\ ($ \pm $ 0.39)} & \makecell{86.51 \\ ($ \pm $ 3.39)} & \makecell{85.23 \\ ($ \pm $ 2.62)} & \makecell{74.92 \\ ($ \pm $ 3.10)} & \makecell{90.39 \\ ($ \pm $ 0.39)} & \makecell{76.47 \\ ($ \pm $ 1.01)} &  \makecell{\textbf{93.80} \\ ($ \pm $ 0.17)} \\
		1 $ \sim $ 8 & 4 & CIFAR10  & \makecell{66.32 \\ ($ \pm $ 0.91)} & \makecell{68.22 \\ ($ \pm $ 0.95)} & \makecell{55.23 \\ ($ \pm $ 0.88)} & \makecell{71.27 \\ ($ \pm $ 0.49)} & \makecell{67.98 \\ ($ \pm $ 0.43)} & \makecell{51.45 \\ ($ \pm $ 1.13)} &  \makecell{\textbf{79.28} \\ ($ \pm $ 0.33)} \\
		\bottomrule[1pt]
	\end{tabular}
\end{table*}

\begin{table*}[!htbp]
	\renewcommand{\arraystretch}{1}
	\caption{Test classification accuracy (mean $ \pm $ std $ \% $) of 5 trials on various datasets and approaches.}
	\label{table_3}
	\centering
	\begin{tabular}{c|c|c|c|c|c|c|c|c|c}%c表示文本居中，c的个数是列数
		\toprule[1pt] 
		class & size & Dataset  & LOG & EXP & LW & RC & CC & PRODEN & SL (our)  \\
		\midrule
		1 $ \sim $ 6 & 3 & MNIST  & \makecell{99.41 \\ ($ \pm $ 0.07)} & \makecell{99.32 \\ ($ \pm $ 0.09)} & \makecell{99.30 \\ ($ \pm $ 0.02)} & \makecell{98.17 \\ ($ \pm $ 0.11)} & \makecell{99.36 \\ ($ \pm $ 0.05)} & \makecell{98.09 \\ ($ \pm $ 0.19)} &  \makecell{\textbf{99.57} \\ ($ \pm $ 0.03)} \\
		1 $ \sim $ 6 & 3 & Fashion  & \makecell{93.25 \\ ($ \pm $ 0.20)} & \makecell{92.69 \\ ($ \pm $ 0.20)} & \makecell{92.95 \\ ($ \pm $ 0.19)} & \makecell{90.72 \\ ($ \pm $ 0.14)} & \makecell{93.44 \\ ($ \pm $ 0.15)} & \makecell{90.62 \\ ($ \pm $ 0.30)} &  \makecell{\textbf{93.99} \\ ($ \pm $ 0.15)} \\
		1 $ \sim $ 6 & 3 & Kuzushiji  & \makecell{91.68 \\ ($ \pm $ 0.58)} & \makecell{89.87 \\ ($ \pm $ 0.64)} & \makecell{89.56 \\ ($ \pm $ 0.38)} & \makecell{84.82 \\ ($ \pm $ 0.44)} & \makecell{91.60 \\ ($ \pm $ 0.41)} & \makecell{83.37 \\ ($ \pm $ 0.64)}  &  \makecell{\textbf{94.29} \\ ($ \pm $ 0.18)} \\
		1 $ \sim $ 6 & 3 & CIFAR10  & \makecell{69.30 \\ ($ \pm $ 1.12)} & \makecell{70.10 \\ ($ \pm $ 0.77)} & \makecell{54.16 \\ ($ \pm $ 3.01)} & \makecell{72.45 \\ ($ \pm $ 0.58)} & \makecell{70.27 \\ ($ \pm $ 0.77)} & \makecell{55.54 \\ ($ \pm $ 0.79)} &  \makecell{\textbf{78.88} \\ ($ \pm $ 0.65)} \\
		\midrule
		4 $ \sim $ 9 & 3 & MNIST  & \makecell{98.62 \\ ($ \pm $ 0.13)} & \makecell{98.32 \\ ($ \pm $ 0.15)} & \makecell{98.43 \\ ($ \pm $ 0.07)} & \makecell{96.53 \\ ($ \pm $ 0.09)} & \makecell{98.63 \\ ($ \pm $ 0.09)} & \makecell{96.18 \\ ($ \pm $ 0.29)} &  \makecell{\textbf{98.99} \\ ($ \pm $ 0.58)} \\
		4 $ \sim $ 9 & 3 & Fashion  & \makecell{94.77 \\ ($ \pm $ 0.18)} & \makecell{94.13 \\ ($ \pm $ 0.12)} & \makecell{94.40 \\ ($ \pm $ 0.16)} & \makecell{91.82 \\ ($ \pm $ 0.39)} & \makecell{94.88 \\ ($ \pm $ 0.11)} & \makecell{91.33 \\ ($ \pm $ 0.36)} &  \makecell{\textbf{95.26} \\ ($ \pm $ 0.19)} \\
		4 $ \sim $ 9 & 3 & Kuzushiji  & \makecell{92.82 \\ ($ \pm $ 0.23)} & \makecell{90.97 \\ ($ \pm $ 0.45)} & \makecell{89.51 \\ ($ \pm $ 2.26)} & \makecell{81.70 \\ ($ \pm $ 0.95)} & \makecell{92.73 \\ ($ \pm $ 0.40)} & \makecell{81.40 \\ ($ \pm $ 1.63)} &  \makecell{\textbf{95.04} \\ ($ \pm $ 0.10)} \\
		4 $ \sim $ 9 & 3 & CIFAR10  & \makecell{82.30 \\ ($ \pm $ 0.82)} & \makecell{80.90 \\ ($ \pm $ 0.86)} & \makecell{67.69 \\ ($ \pm $ 0.31)} & \makecell{84.88 \\ ($ \pm $ 0.42)} & \makecell{81.62 \\ ($ \pm $ 0.51)} & \makecell{65.47 \\ ($ \pm $ 1.85)} &  \makecell{\textbf{87.79} \\ ($ \pm $ 0.35)} \\
		\bottomrule[1pt]
	\end{tabular}
\end{table*}

\begin{table*}[!htbp]
	\renewcommand{\arraystretch}{1}
	\caption{Test classification accuracy (mean $ \pm $ std $ \% $) of 5 trials on various dataset and approaches.}
	\label{table_3}
	\centering
	\begin{tabular}{c|c|c|c|c|c|c|c|c|c}%c表示文本居中，c的个数是列数
		\toprule[1pt] 
		class & size & Dataset  & LOG & EXP & LW & RC & CC & PRODEN & SL (our)  \\
		\midrule
		1 $ \sim $ 4 & 2 & MNIST  & \makecell{99.53 \\ ($ \pm $ 0.06)} & \makecell{99.41 \\ ($ \pm $ 0.07)} & \makecell{99.50 \\ ($ \pm $ 0.11)} & \makecell{98.79 \\ ($ \pm $ 0.06)} & \makecell{99.57 \\ ($ \pm $ 0.08)} & \makecell{98.74 \\ ($ \pm $ 0.12)} &  \makecell{\textbf{99.68} \\ ($ \pm $ 0.11)} \\
		1 $ \sim $ 4 & 2 & Fashion  & \makecell{95.60 \\ ($ \pm $ 0.04)} & \makecell{94.93 \\ ($ \pm $ 0.06)} & \makecell{95.17 \\ ($ \pm $ 0.13)} & \makecell{94.28 \\ ($ \pm $ 0.13)} & \makecell{95.72 \\ ($ \pm $ 0.05)} & \makecell{93.99 \\ ($ \pm $ 0.11)} &  \makecell{\textbf{96.07} \\ ($ \pm $ 0.10)} \\
		1 $ \sim $ 4 & 2 & Kuzushiji  & \makecell{95.13 \\ ($ \pm $ 0.17)} & \makecell{94.64 \\ ($ \pm $ 0.28)} & \makecell{94.73 \\ ($ \pm $ 0.25)} & \makecell{91.67 \\ ($ \pm $ 0.49)} & \makecell{95.09 \\ ($ \pm $ 0.20)} & \makecell{91.37 \\ ($ \pm $ 0.65)}  &  \makecell{\textbf{96.00} \\ ($ \pm $ 0.17)} \\
		1 $ \sim $ 4 & 2 & CIFAR10  & \makecell{80.76 \\ ($ \pm $ 1.12)} & \makecell{81.85 \\ ($ \pm $ 0.43)} & \makecell{80.92 \\ ($ \pm $ 0.87)} & \makecell{82.43 \\ ($ \pm $ 0.78)} & \makecell{81.45 \\ ($ \pm $ 0.30)} & \makecell{70.38 \\ ($ \pm $ 1.02)} &  \makecell{\textbf{85.42} \\ ($ \pm $ 0.56)} \\
		\midrule
		4 $ \sim $ 7 & 2 & MNIST  & \makecell{99.29 \\ ($ \pm $ 0.09)} & \makecell{99.19 \\ ($ \pm $ 0.06)} & \makecell{99.28 \\ ($ \pm $ 0.09)} & \makecell{98.25 \\ ($ \pm $ 0.13)} & \makecell{99.26 \\ ($ \pm $ 0.10)} & \makecell{98.15 \\ ($ \pm $ 0.18)} &  \makecell{\textbf{99.58} \\ ($ \pm $ 0.04)} \\
		4 $ \sim $ 7 & 2 & Fashion  & \makecell{92.52 \\ ($ \pm $ 0.25)} & \makecell{91.63 \\ ($ \pm $ 0.29)} & \makecell{92.17 \\ ($ \pm $ 0.28)} & \makecell{90.73 \\ ($ \pm $ 0.17)} & \makecell{92.72 \\ ($ \pm $ 0.12)} & \makecell{90.03 \\ ($ \pm $ 0.30)} &  \makecell{\textbf{93.26} \\ ($ \pm $ 0.12)} \\
		4 $ \sim $ 7 & 2 & Kuzushiji  & \makecell{95.33 \\ ($ \pm $ 0.13)} & \makecell{94.23 \\ ($ \pm $ 0.40)} & \makecell{94.00 \\ ($ \pm $ 0.20)} & \makecell{90.91 \\ ($ \pm $ 0.31)} & \makecell{95.17 \\ ($ \pm $ 0.22)} & \makecell{90.38 \\ ($ \pm $ 0.41)} &  \makecell{\textbf{96.44} \\ ($ \pm $ 0.13)} \\
		4 $ \sim $ 7 & 2 & CIFAR10  & \makecell{67.88 \\ ($ \pm $ 1.83)} & \makecell{71.53 \\ ($ \pm $ 0.40)} & \makecell{69.13 \\ ($ \pm $ 0.47)} & \makecell{72.48 \\ ($ \pm $ 0.78)} & \makecell{71.50 \\ ($ \pm $ 1.12)} & \makecell{56.04 \\ ($ \pm $ 1.92)} &  \makecell{\textbf{74.73} \\ ($ \pm $ 0.91)} \\
		\midrule
		7 $ \sim $ 10 & 2 & MNIST  & \makecell{98.82 \\ ($ \pm $ 0.12)} & \makecell{98.63 \\ ($ \pm $ 0.11)} & \makecell{98.73 \\ ($ \pm $ 0.13)} & \makecell{97.77 \\ ($ \pm $ 0.20)} & \makecell{98.82 \\ ($ \pm $ 0.10)} & \makecell{97.51 \\ ($ \pm $ 0.21)} &  \makecell{\textbf{99.26} \\ ($ \pm $ 0.06)} \\
		7 $ \sim $ 10 & 2 & Fashion  & \makecell{97.50 \\ ($ \pm $ 0.16)} & \makecell{97.22 \\ ($ \pm $ 0.13)} & \makecell{97.45 \\ ($ \pm $ 0.13)} & \makecell{96.75 \\ ($ \pm $ 0.07)} & \makecell{97.66 \\ ($ \pm $ 0.10)} & \makecell{96.77 \\ ($ \pm $ 0.14)} &  \makecell{\textbf{97.92} \\ ($ \pm $ 0.08)} \\
		7 $ \sim $ 10 & 2 & Kuzushiji  & \makecell{95.46 \\ ($ \pm $ 0.36)} & \makecell{94.70 \\ ($ \pm $ 0.35)} & \makecell{93.39 \\ ($ \pm $ 0.57)} & \makecell{88.07 \\ ($ \pm $ 0.59)} & \makecell{95.48 \\ ($ \pm $ 0.35)} & \makecell{88.46 \\ ($ \pm $ 0.49)} &  \makecell{\textbf{97.39} \\ ($ \pm $ 0.17)} \\
		7 $ \sim $ 10 & 2 & CIFAR10  & \makecell{91.97 \\ ($ \pm $ 0.67)} & \makecell{91.15 \\ ($ \pm $ 0.62)} & \makecell{91.61 \\ ($ \pm $ 0.50)} & \makecell{94.07 \\ ($ \pm $ 0.27)} & \makecell{92.84 \\ ($ \pm $ 0.64)} & \makecell{79.54 \\ ($ \pm $ 0.09)} &  \makecell{\textbf{94.14} \\ ($ \pm $ 0.09)} \\
		\bottomrule[1pt]
	\end{tabular}
\end{table*}

\section{Experiments}
In this part, we do extensive experiments to evaluate the performance of the proposed approach. We use SL to define our proposed approach and compare its effectiveness to ordinary label learning as well as the state-of-the-art WSL approaches. Each experiment was carried out under the assumption that fairness was satisfied. All experiments are implemented based on PyTorch and run on a NVIDIA GeForce RTX 3090 GPU.
 
\subsection{Experimental Setting}

\textbf{Datasets:} Similar to other datasets used by WSL\citep{cll_1, cll_3, cll_2, cll_5, cll_7, cll_8},  we also used four widely-used benchmark datasets in our experiments, namely MNIST \citep{MNIST}, Fashion \citep{Fashion}, Kuzushiji \citep{Kuzushiji} and CIFAR10 \citep{CIFAR10}. 

We use OVR strategy with square loss $ \psi(z) = (1-z)^2/4 $ to train the classifier. For MNIST, Fashion and Kuzushiji, we use the nerual network which has two convolutional layers and two fully-connected layers for all methods. The number of the epoch was set to 100 and learning rate was selected from $ \{1e-1, 5e-2, 1e-2, 5e-3, 1e-3, 5e-4, 1e-4, 5e-5\} $. For CIFAR10, we employ 32-layer ResNet \citep{Resnet} for all methods. The number of the epoch was set to 200. We used stochastic gradient descent (SGD) \citep{SGD} optimizer with momentum  set to 0.9. Here weight-decay was select from  $ \{1e-6, 1e-5, 1e-4, 1e-3, 1e-2\} $ and the learning rate was selected from $ \{1e-1, 5e-2, 1e-2, 5e-3, 1e-3 \} $.

\textbf{Compared Methods:} We compare the effectiveness of the proposed method to ordinary label approach as well as six state-of-the-art WSL approaches, including two multiple complementary label learning approaches: LOG and EXP \citep{cll_5} and four partial label learning approaches: LW \citep{lw}, RC and CC \citep{rc_cc}, and PRODEN \citep{proden}. In LOG and EXP, \citet{cll_5} treated multiple complementary labels as a whole. In LW, \citet{lw} provided the leverage option for the first time to take into account the trade-off between losses on partial labels and non-partial ones. RC and CC are provably consistent partial label learning methods. PRODEN is a model-independent and loss-independent partial label learning approach.

\subsection{Comparison with weakly labels}
In this experiment, we use all of classes from dataset. Table 1 shows the mean and standard deviation of test classification accuracy on four datasets. Here the size of the labels set for each instance was set to 8, 7, 6, 5, 4, and 3. The proportions of the used training data were 0.2, 0.3, 0.4, and 0.5 respectively. We compare all the methods in the same settings, using training samples with the same amount of data. As can be seen from Table 1, our proposed method has very outstanding performance. Furthermore, when the number of labels of each sample decreases, the performance of all methods has different degrees of degradation, especially the performance of LW, RC, PRODEN on MNIST, Fashion, Kuzushiji. These methods are difficult to maintain excellent performance in extreme cases, but our method can still adapt well to this situation,  which confirms the superiority of the proposed approach. Figure 2 shows the accuracy of each epoch on the test set of MNIST and CIFAR10 datasets for all methods. And the results of Fashion and Kuzushiji are available in Appendix. As can be observed, the proposed approach exhibits superior functionality and more stability.

\subsection{Comparison with ordinary labels}
On these benchmark datasets, we compare the effects of using all ordinary labels and using stochastic labels. For stochastic labels, we set different criteria. The size of stochastic labels $ \tilde{Y} $ here was set to 8, 7, 6, and 5. The proportions of the training data corresponding to the stochastic labels that did not contain true labels were 0.2, 0.3, 0.4, and 0.5, respectively. 
In Table 2, we report the mean and standard deviation of test classification accuracy on four datasets. Table 2 shows the performance comparison between ordinary labels(OL) and stochastic labels(SL). Here, OL represents learning from ordinary labels. From Table 2, we can see that the less informative stochastic labels have better performance than using all ordinary labels. We believe that assigning a ordinary label to those data that are more challenging to classify leads to overfitting. By giving the model some weakly labels to choose from on its own, it might be easier for it to fit the data. This also inspired us to study what kind of label is needed in the future.

\subsection{Comparison on  smaller scale with half of labels}
In order to test the performance of stochastic labels on certain specific classes, we divided the entire candidate labels set into smaller scale. In this experiment, the smaller scale of the division mainly includes the following three case: 1) 8 classes with half of labels; 2) 6 classes with half of labels; 3) 4 classes with half of labels. Table 3 $ \sim $ 5 reports the mean and standard deviation of test classification accuracy of these case on benchmark datasets. As can be observed, on a smaller scale, all methods have some performance gains due to the relative increase in the number of labels provided. On the other hand, all methods have a significant performance degradation under the 1-6, 4-7 division, although our method still maintains the best performance.

\section{Conclusion}
In this paper, we proposed a novel labeling mechanism called \emph{stochastic label}. For each instance, a crowdsourced worker only needs to identify the correct class label from the limited number of randomly selected labels or annotate None, which significantly reduces the time cost of labeling task. We showed that an unbiased risk estimator for the classification risk can be obtained using these stochastically labeled data. We demonstrated that the proposed approach obtains the optimal parametric rate and theoretically established estimation error bounds for it. Finally, we experimentally demonstrated the superiority of the proposed approach.
 
Note that the labels were chosen at random, which also appears in other weakly supervised learning approaches. It would be interesting to take the initiative to choose some stochastic labels that can improve the performance of model.

\bibliography{example_paper}
\bibliographystyle{icml2023}

\end{document}